\documentclass[journal]{IEEEtran}
\usepackage{stfloats}
\usepackage{url}
\usepackage{verbatim}
\usepackage{balance}
\usepackage{amsmath,amssymb,amsfonts}
\usepackage{graphicx}
\usepackage{textcomp}
\usepackage{xcolor}
\usepackage{makecell}
\usepackage{gensymb}
\usepackage{multirow}
\usepackage{amssymb}% http://ctan.org/pkg/amssymb
\usepackage{pifont}% http://ctan.org/pkg/pifont
\usepackage{subcaption}
\usepackage{algorithm}
\usepackage{algpseudocode}
\usepackage{hyperref}
\usepackage{array}
\usepackage{adjustbox}

\def\BibTeX{{\rm B\kern-.05em{\sc i\kern-.025em b}\kern-.08em
		T\kern-.1667em\lower.7ex\hbox{E}\kern-.125emX}}

\newcolumntype{?}{!{\vrule width 1pt}}

\usepackage{tikz,xcolor,hyperref}
\definecolor{lime}{HTML}{A6CE39}
\DeclareRobustCommand{\orcidicon}{
	\begin{tikzpicture}
		\draw[lime, fill=lime] (0,0) 
		circle [radius=0.16] 
		node[white] {{\fontfamily{qag}\selectfont \tiny ID}};
		\draw[white, fill=white] (-0.0625,0.095) 
		circle [radius=0.007];
	\end{tikzpicture}
	\hspace{-2mm}
}

\foreach \x in {A,B}{\expandafter\xdef\csname orcid\x\endcsname{\noexpand\href{https://orcid.org/\csname orcidauthor\x\endcsname}
		{\noexpand\orcidicon}}
}

\newcommand\copyrighttext{%
	\footnotesize © 2023 IEEE. Personal use of this material is permitted. Permission from IEEE must be 
	obtained for all other uses, in any current or future media, including 
	reprinting/republishing this material for advertising or promotional purposes, creating new 
	collective works, for resale or redistribution to servers or lists, or reuse of any copyrighted 
	component of this work in other works. \href{dx.doi.org/10.1109/MITS.2023.3268032}{DOI:10.1109/MITS.2023.3268032} }
\newcommand\copyrightnotice{%
	\begin{tikzpicture}[remember picture,overlay]
		\node[anchor=south,yshift=10pt] at (current page.south) {\fbox{\parbox{\dimexpr\textwidth-\fboxsep-\fboxrule\relax}{\copyrighttext}}};
	\end{tikzpicture}%
}

\begin{document}
	\title{Uncertainty-Aware Vehicle Energy Efficiency Prediction using an Ensemble of Neural Networks}

	% author names and affiliations
	\author{Jihed Khiari  \orcidA{} \IEEEmembership {Student Member, IEEE} and Cristina~Olaverri-Monreal \orcidB{}~\IEEEmembership{Senior Member, IEEE}% <-this stops a space
		
		\thanks{Jihed Khiari and Cristina Olaverri-Monreal are with ITS-Chair for Sustainable Transport Logistics 4.0., Johannes Kepler University, Linz in Austria. e-mail: (jihed.khiari, cristina.olaverri-monreal)@jku.at}}
	
	%\markboth{}
	
	\maketitle
	
	\copyrightnotice
	
	\begin{abstract}
		The transportation sector accounts for about 25\% of global greenhouse gas emissions. Therefore, an improvement of energy efficiency in the traffic sector is crucial to reducing the carbon footprint. Efficiency is typically measured in terms of energy use per traveled distance, e.g. liters of fuel per kilometer. Leading factors that impact the energy efficiency are the type of vehicle, environment, driver behavior, and weather conditions. These varying factors introduce uncertainty in estimating the vehicles' energy efficiency. We propose in this paper an ensemble learning approach based on deep neural networks (ENN) that is designed to reduce the predictive uncertainty and to output measures of such uncertainty. We evaluated it using the publicly available Vehicle Energy Dataset (VED) and compared it with several baselines per vehicle and energy type. The results showed a high predictive performance and they allowed to output a measure of predictive uncertainty.
		
	\end{abstract}
	

	\section{Introduction}
	\label{sec:introduction}
	The growth of e-commerce has contributed to an increase in the transportation systems' dependency on burning fossil fuels such as gasoline and
	diesel~\cite{liu2020mobile}. In general, transportation is estimated to generate a quarter of the total greenhouse gas emissions~\cite{owidco2andgreenhousegasemissions}. In this context, key alternative energy sources of transportation such as battery-equipped vehicles (BEV) can improve energy efficiency and reduce CO2 emissions in the traffic sector~\cite{validi2021analysis}~\cite{yadlapalli2022review}. As a consequence, a real-world close evaluation and research towards a sustainable transport strategy is crucial to reduce energy consumption~\cite{gonccalves2014smartphone}.
	While energy consumption is a useful metric for comparing the total energy used by different vehicles or modes of transportation, it does not provide a complete picture of a vehicle's energy performance. Energy efficiency is a measure of how effectively a vehicle uses energy to perform a specific task, such as moving a certain distance or carrying a certain load. It takes into account not only the amount of energy consumed, but also the work achieved by the vehicle in relation to that energy consumption. For example, two vehicles may consume the same amount of energy, but one may be more efficient at converting that energy into motion, resulting in better performance and lower energy costs.

	Different factors such as: vehicle type, size, and type of fuel, use of electric energy, the outside temperature, as well as the type of road (highway vs. urban area) can affect the use of energy~\cite{vaezipour2015reviewing}~\cite{lu2010analysis}. By estimating the energy efficiency under given circumstances, individual and commercial vehicle users can make more informed decisions about their trip planning and better understand their impact. For instance, given a fleet of vehicles that can be used for deliveries, an operator can plan the trips according to the estimated energy efficiency. Furthermore, private car users can benefit from historical information and predictions of the energy efficiency for future trips. For instance, predictions of energy efficiency can inform the choice of route or timing of a planned trip. In both cases, the higher the energy efficiency, the lower the costs and the environmental impact. 
	
	On the other hand, traditional prediction methods can be highly sensitive and inaccurate as they are affected by factors that affect the energy consumption of a vehicle. Furthermore, prior knowledge about energy efficiency can diverge from reality, as the vehicles can be utilized in various contexts other than the ones tested by the manufacturer. In~\cite{kwon2006determinants}, the authors report about a 10\% gap between the official fuel consumption rate of new cars and on-road fuel consumption rate, which generally corroborates the findings of earlier studies such as~\cite{schipper1994new} and~\cite{sorrell1992fuel}.  It is therefore useful to generate estimations based on commonly available ground-truth data that reflect specificities of the vehicles' usage: environment, weather conditions, driver behavior, etc. 
	
	To tackle this challenge, we propose in this paper a machine learning approach consisting in an ensemble of neural networks (ENN) that predicts energy efficiency and estimates the uncertainty in the prediction. By using a diverse set of base learners, ensembles are able to deliver results that are less sensitive to variations in the data. Furthermore, we propose an ensemble that outputs predictions, as well as a measure of confidence. 
	
	We investigate in this paper the use of an ensemble, the base learners and scoring functions. Then, we validate our model on the Vehicle Energy Dataset (VED)~\cite{oh2020vehicle} that is an openly available dataset spanning over a year and including trips undertaken by several types of vehicles. 
	
	Unlike other works in the literature, our approach does not rely on complex kinetic, chemical, or physical models, nor does it require prior knowledge about the energy efficiency measures provided by the vehicle manufacturer. Furthermore, it is not aimed at a specific type of vehicle. This results in a practical approach, which can be adapted to the dataset at hand and then be used to improve energy efficiency predictions. 
	
	Our main contributions are:
	\begin{enumerate}
		\item Proposing an ensemble of deep neural networks (ENN) that can output energy efficiency predictions for different types of vehicles that utilize fuel and/or an electric battery.
		\item Proposing an ensemble that reduces predictive uncertainty and outputs predictions as well as a measure of uncertainty. 
		\item Presenting an extensive evaluation of the approach per type of vehicle and type of energy, as well as a comparison with several baselines. 
	\end{enumerate}

	The remainder of this paper is structured as follows. The next section considers related work in the field of predictive accuracy and uncertainty estimation for vehicle energy efficiency. Section~\ref{sec:method} describes the steps that were
	followed to implement our approach. Section~\ref{sec:results} presents the results regarding the predicted energy values. Finally, Sections~\ref{sec:discussion} and~\ref{sec:conclusion} discuss the findings and conclude the paper.

	\section{Related Work}
	\label{sec:related}
	
	Most available works in the literature about energy efficiency rely either on manufacturer data (which announce an estimated measure of efficiency for a given vehicle model), or complex kinetic, physical, or chemical models that estimate the energy consumption of a given vehicle based on detailed data. Furthermore, such methods typically focus on only one type of vehicle. On the other hand, there are many studies that tackle how vehicle design and manufacturing choices can enable a higher energy efficiency such as~\cite{sajadi2018ecological}~\cite{koehler2017energy}~\cite{lu2019energy} which respectively focus on the advanced driver assistance system, the torque vectoring control, and the adaptive cruise control. Other studies such as~\cite{dong2022predictive} investigate an energy-efficient driving strategy in the context of connected vehicles.
	
	We propose to predict the energy efficiency for different types of vehicles based solely on floating car data that includes measures of instantaneous energy consumption and/or measures that can allow to estimate the energy consumption.  
	
	Focusing on trip efficiency estimation, there are mainly three types of approaches that have been followed in the literature:
	
	\begin{enumerate}
		\item Approaches that rely on mechanical, dynamic, and kinetic car data to derive models for energy consumption estimation, such as~\cite{song2009estimation}, ~\cite{miri2021electric}, and~\cite{iclodean2017comparison}.
		\item Approaches that use a data-driven approach, where energy data and/or a measure of energy efficiency is available, such as~\cite{yin2015fuel}, ~\cite{syahputra2016application}, ~\cite{thomas2009fuel}, and~\cite{ahn2002estimating}.
		\item Approaches that combine the first two types of approaches. For instance,~\cite{cappiello2002statistical} combines statistical modeling with load-based models which simulate the physical phenomena that generate emissions. 
	\end{enumerate}
	
	In~\cite{song2009estimation}, the authors used vehicle-specific power to estimate fuel efficiency per time and per distance. Vehicle-specific power (VSP) is a formalism used in the evaluation of vehicle emissions. It was first proposed in~\cite{jimenez1998understanding}. Informally, it is the sum of the loads resulting from aerodynamic drag, acceleration, rolling resistance, and hill climbing, all divided by the mass of the vehicle. Conventionally, it is reported in kilowatts per tonne, the instantaneous power demand of the vehicle divided by its mass. VSP, combined with dynamometer and remote-sensing measurements, can be used to determine vehicle emissions and, by extension, fuel consumption. The  authors of~\cite{wu2011development} proposed a backpropagation network to predict fuel consumption. They relied on a dataset from an auto energy website in Taiwan which included fuel data, and further important factors for fuel consumption such as: car manufacturer, engine styles, vehicle's weight, vehicle type, and transmission types. Other common approaches comprise analyzing datasets that include fuel efficiency information expressed in terms of mileage per gallon or equivalently fuel per unit of traveled distance, such as in~\cite{yin2015fuel} and~\cite{syahputra2016application}. Given the fuel efficiency data, the authors of~\cite{yin2015fuel} used several supervised learning methods to predict the fuel efficiency. A fuzzy inference method to predict mileage per gallon was detailed in~\cite{syahputra2016application}.
	
	Energy efficiency prediction can be considered as one of the measures that can help with delivery trip optimization, alongside other considerations, such as: travel time prediction~\cite{lin2005review}~\cite{khiari2020boosting}, route optimization~\cite{ichoua2007planned}~\cite{archetti2008optimization}, and order assignment~\cite{liu2018time}~\cite{chu2021data}.
	
	%Nevertheless, energy efficiency can be highly affected by different factors, thus resulting in high uncertainty. 
	Among the data-driven approaches for estimating efficiency, we can consider deep neural networks. Deep neural networks (DNN) are prominent machine learning methods that have achieved considerable successes in many fields such as computer vision, natural language processing, and robotics~\cite{lecun2015deep}~\cite{zhang2018survey}~\cite{sunderhauf2018limits}. However, they can be prone to outputting wrong results with high confidence. Such high confidence can present security risks~\cite{amodei2016concrete}~\cite{carlini2017towards}.
	Adversarial training \cite{kurakin2016adversarial} is a method that can help tackle such issue by increasing robustness. As for quantifying uncertainty, the most common approach is to use Bayesian networks~\cite{bernardo2009bayesian}, which are however complex to implement and computationally expensive. Therefore, simpler methods such as the DNN ensemble described in~\cite{lakshminarayanan2017simple} can be useful and easily deployed.
	
	The use of ensembles to improve predictive performance is common~\cite{dietterich2000ensemble}~\cite{mendes2012ensemble}. In~\cite{lakshminarayanan2017simple}, the authors described how to use an ensemble to quantify predictive uncertainty. To the authors' knowledge, a similar method has not been used in the context of energy efficiency prediction for different types of vehicles. We therefore study in this work the usability of such an ensemble for providing uncertainty-aware predictions. 
	
	\section{Methodology}
	\label{sec:method}

	\subsection{Problem Formulation}
\begin{figure*}
		\centering
		\begin{subfigure}[t]{0.35\textwidth}
			\centering
			\includegraphics[width=\textwidth]{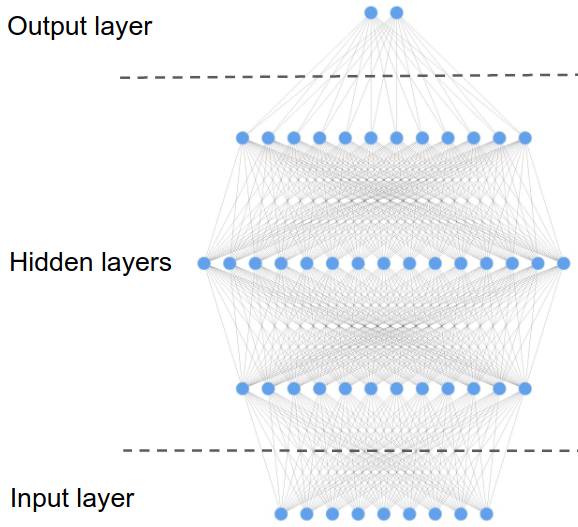}
			\caption{Neural network architecture}
			\label{fig:nn_arch}
		\end{subfigure}
		\hfill
		\begin{subfigure}[t]{0.5\textwidth}
			\centering
			\includegraphics[width=\textwidth]{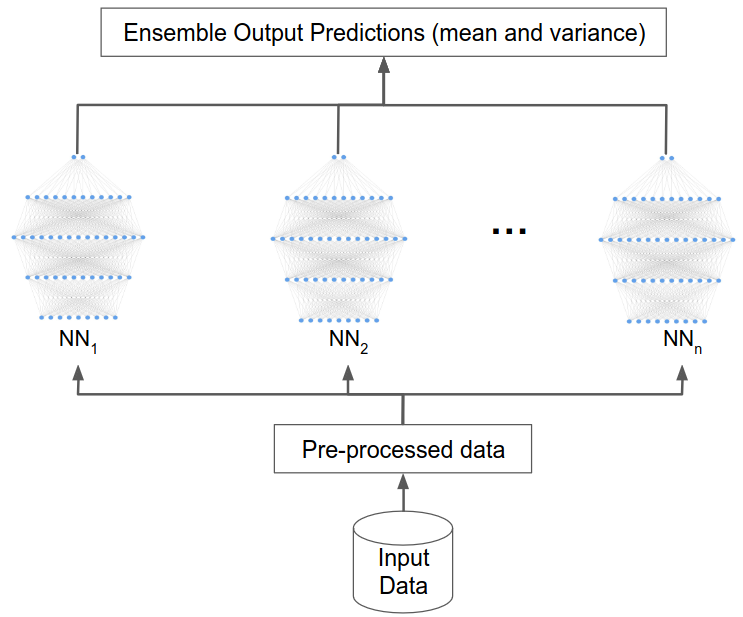}
			\caption{Ensemble framework}
			\label{fig:framework}
		\end{subfigure}
		\caption{Ensemble and neural network representations}
		\label{fig:representations}
	\end{figure*}

	As input, we used vehicle data including energy information. We then divided the data per vehicle type. After processing, we generated relevant features and computed the energy efficiency per trip for each type of vehicle. We proceeded to train an ensemble of neural networks. The output consisted in energy efficiency predictions for each trip in the testing set. Figure~\ref{fig:framework} summarizes the process.
	
	\subsection{Data Description}
	
	We relied in this study on the Vehicle Energy Dataset (VED) that is openly available and described in~\cite{oh2020vehicle}. The dataset is divided into dynamic data and static data. The dynamic data capture GPS trajectories of vehicles along with their time series data of fuel, energy, speed, and auxiliary power usage. The data were collected, with a granularity of one second through onboard OBD-II loggers for the period of November 2017 to November 2018. The data does not include information identifying the driver.
	Tables~\ref{Table: Dataset} and~\ref{Table: Attributes} show a brief summary of the dataset key aspects and attributes.
	
	The static data captured the constant information about the vehicles, such as type, class, and weight. The list of  static attributes is shown in Table~\ref{Table: Static} along with example values. By combining the static and dynamic data, we were able to uniquely identify the trips and the vehicles in the dataset, obtaining a  fleet that consisted of a total of 383 personal cars in Ann Arbor, Michigan, USA.
	We further classified the cars into 264 gasoline vehicles (ICE), 92 hybrid electric vehicles (HEV), and 27 plug-in hybrid electric (PHEV) or electric vehicles (EV) according to their energy source: 
	
	\begin{figure}
		\centering
		\includegraphics[width=0.33\textwidth]{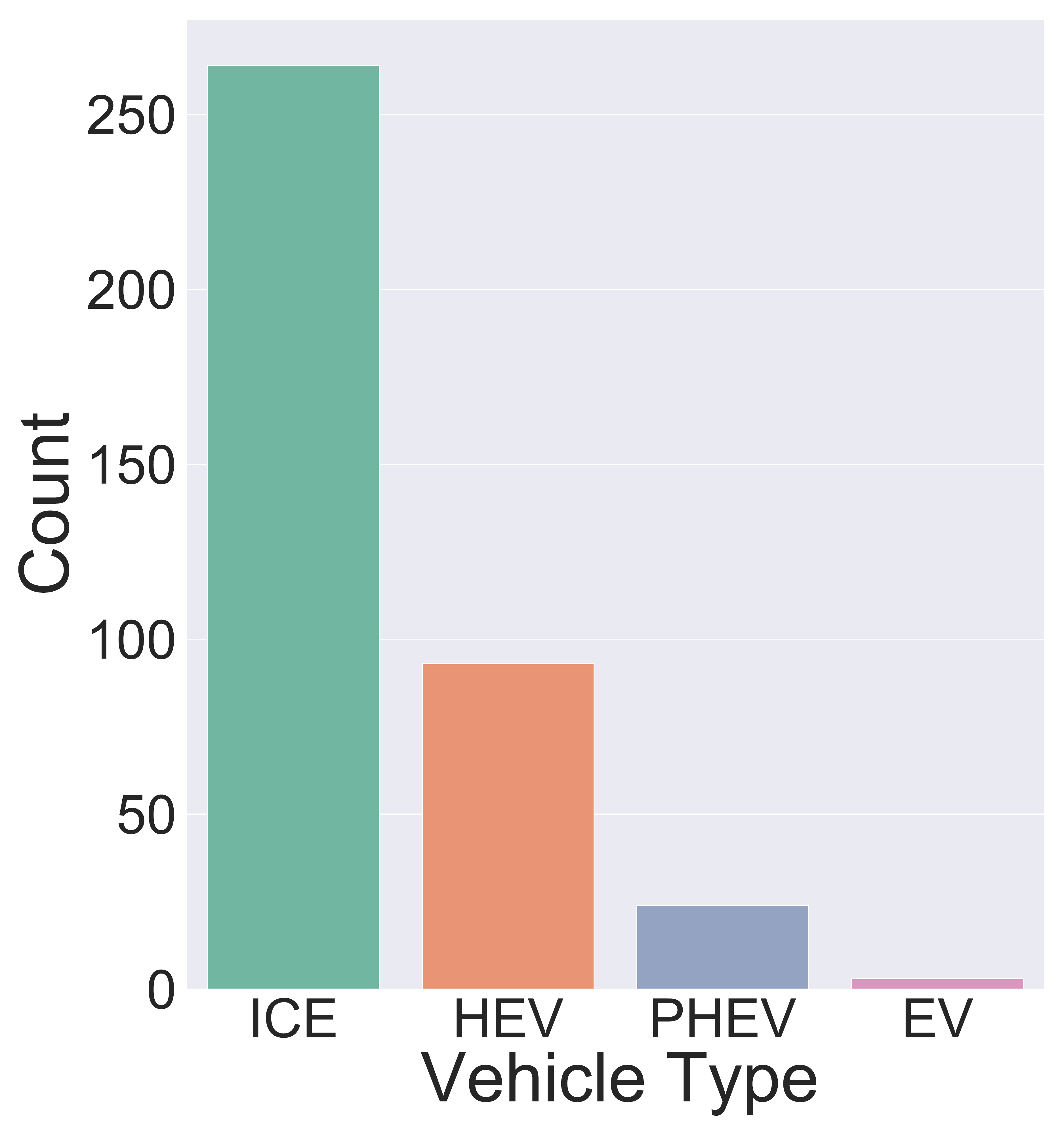}
		\caption{Vehicles distribution per engine type}
		\label{fig:vehicle_types}
	\end{figure}

	\begin{enumerate}
		\item \textbf{ICE:} Internal combustion engine
		\item \textbf{PHEV:} Plug-in hybrid electric vehicles
		\item \textbf{HEV:} Hybrid electric vehicles
		\item \textbf{EV:} Electric vehicles
	\end{enumerate}
	
\begin{figure*}
		\centering
		
		%4*1
		%   \begin{subfigure}{0.24\textwidth}
			%       \includegraphics[width=\textwidth]{ice_dur1}
			%       \caption{ICE}
			%       \label{fig:ice}
			%   \end{subfigure}
		%   %\hfill
		%   \begin{subfigure}{0.24\textwidth}
			%       \includegraphics[width=\textwidth]{phev_dur1}
			%       \caption{PHEV}
			%       \label{fig:phev}
			%   \end{subfigure}
		%   %   %\hfill
		%   %\vskip\baselineskip
		%   \begin{subfigure}{0.24\textwidth}
			%       \includegraphics[width=\textwidth]{hev_dur1}
			%       \caption{HEV}
			%       \label{fig:hev}
			%   \end{subfigure}
		%   %\hfill
		%   \begin{subfigure}{0.24\textwidth}
			%       \includegraphics[width=\textwidth]{ev_dur1}
			%       \caption{EV}
			%       \label{fig:ev}
			%   \end{subfigure} 

		%2*2
		\begin{subfigure}{0.23\textwidth}
			\includegraphics[width=\textwidth]{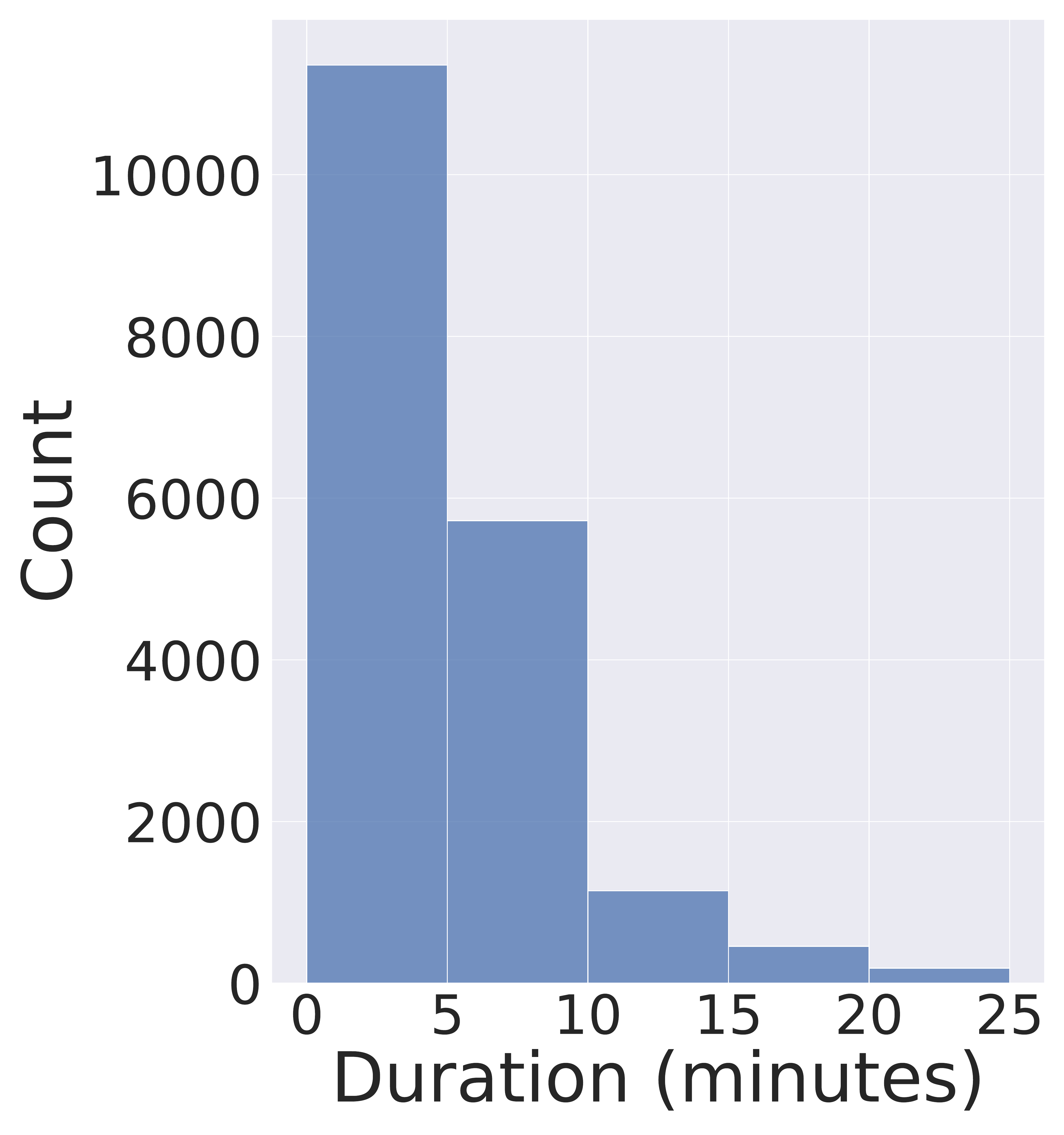}
			\caption{ICE}
			\label{fig:ice}
		\end{subfigure}
		%\hfill
		\begin{subfigure}{0.23\textwidth}
			\includegraphics[width=\textwidth]{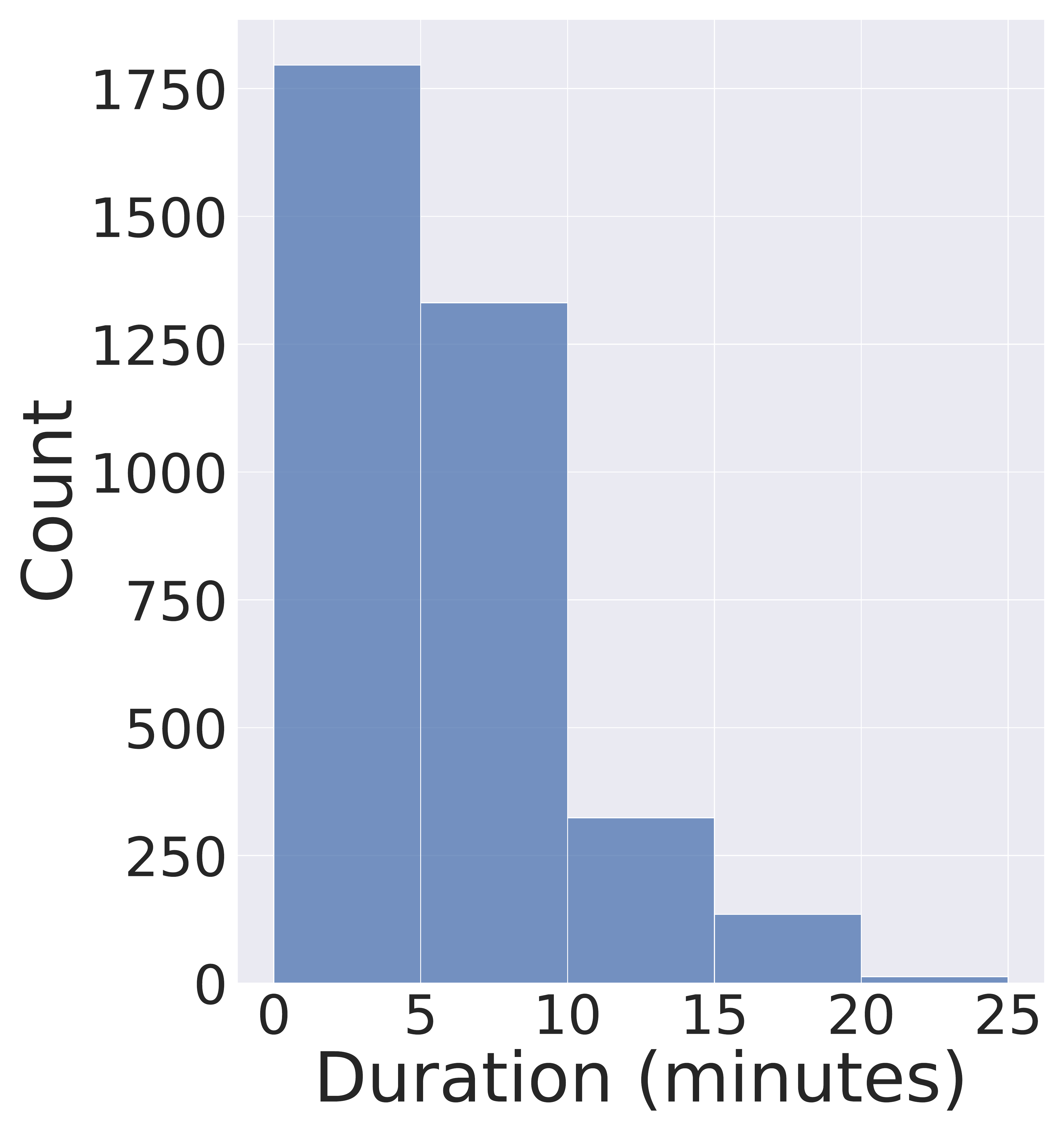}
			\caption{PHEV}
			\label{fig:phev}
		\end{subfigure}
		%   %\hfill
		%\vskip\baselineskip
		\begin{subfigure}{0.23\textwidth}
			\includegraphics[width=\textwidth]{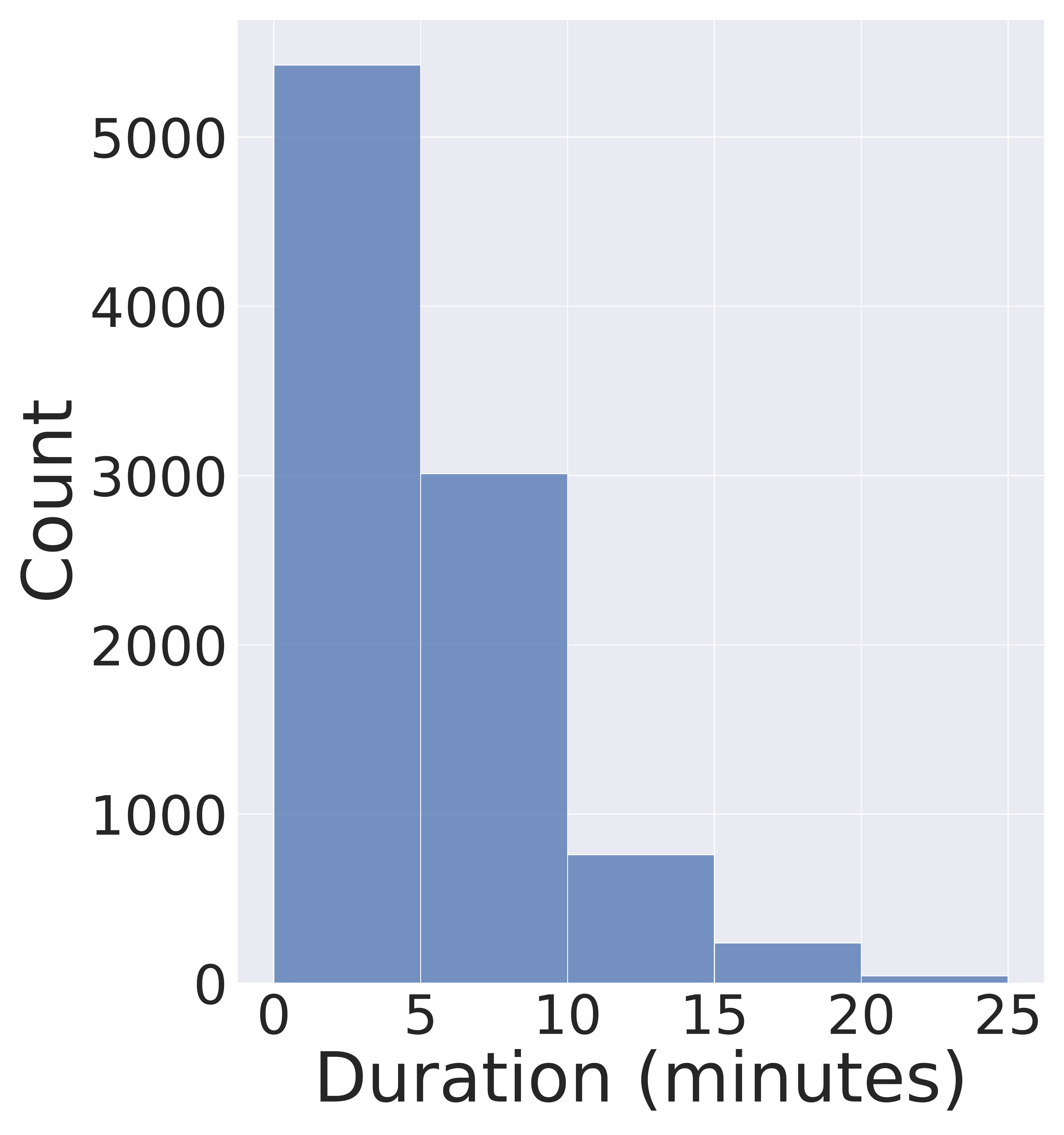}
			\caption{HEV}
			\label{fig:hev}
		\end{subfigure}
		%\hfill
		\begin{subfigure}{0.23\textwidth}
			\includegraphics[width=\textwidth]{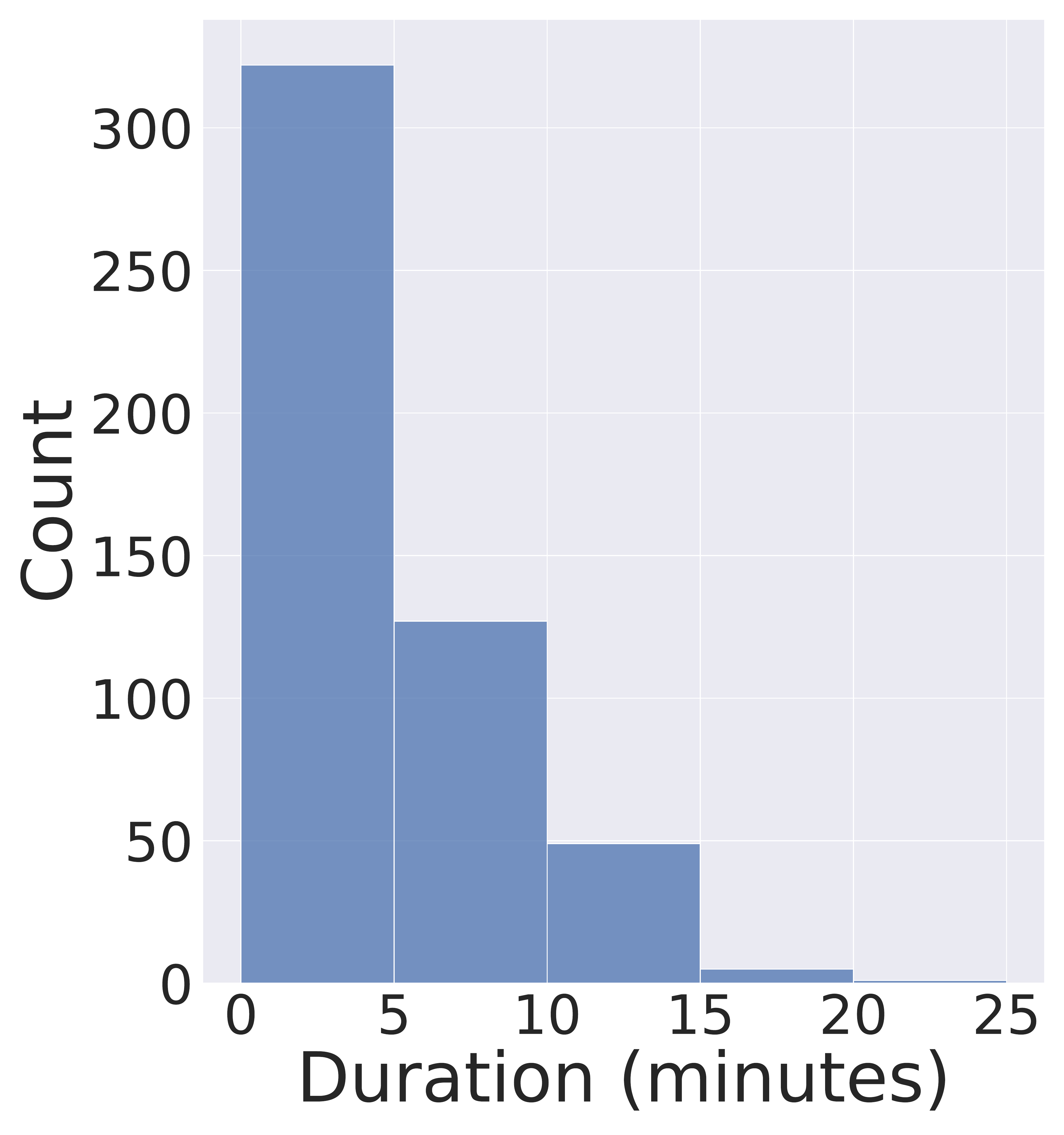}
			\caption{EV}
			\label{fig:ev}
		\end{subfigure} 
		
		\caption{Histograms of trip durations in minutes per vehicle type} 
		\label{fig:trip_durations}
	\end{figure*}

	The distribution of vehicles per engine types is depicted in Figure~\ref{fig:vehicle_types}. The trips show spatio-temporal diversity: they take place at different times of the day throughout the year, and they are distributed across different types of roads in highways and urban areas. The type of road is however not encoded in the dataset as an attribute. Figure~\ref{fig:trip_durations} depicts the distribution of trip durations for each vehicle type. We note that for all vehicle types, the durations of trips are majoritarily under 10 minutes. For ICE vehicles, the histogram shows a considerably higher trip counts than other vehicle types.

	\begin{table}[]
		\centering
		\caption{Datasets summary.}
		\label{Table: Dataset}
		\begin{tabular}{ l l l } 
			\Xhline{2\arrayrulewidth}
			\textbf{Features} & \textbf{VED}  \\ 
			\Xhline{2\arrayrulewidth}
			
			Number of Vehicles  & 264  \\ 
			\hline
			Number of Trips & 18963 \\ 
			\hline
			Traveled distance (km) & 320792  \\ 
			\hline
			Average Number of Trips per Day  & 53 \\ 
			\hline
			Average Trip Duration (minutes) & 15  \\
			\hline
			Location & Ann Arbor, USA \\ 
			\hline
			Time Period  & 1 Year \\  
			\Xhline{2\arrayrulewidth}   
		\end{tabular}

	\end{table}

	\begin{table}[]
		\centering
		\caption{Dynamic Data Attributes per Dataset}
		\label{Table: Attributes}
		\begin{tabular}{l l l }
			\Xhline{2\arrayrulewidth}
			& & \textbf{Attributes}\\ 
			\Xhline{2\arrayrulewidth}
			\textbf{Time} & \multicolumn{2}{c}{Timestamp}  \\
			\hline
			\textbf{GPS} & \multicolumn{2}{c}{Latitude/Longitude (deg)} \\
			\hline
			&\multicolumn{2}{c}{Vehicle Speed (km/h)} \\
			\hline
			\multirow{7}{*}{\textbf{Engine Info}} & \multicolumn{2}{c}{Engine RPM (rev/min)}  \\ \cline{2-3}
			
			& \multirow{3}{*}{\textbf{Fuel Info}} & Mass Air Flow (g/s) \\ 
			
			& & Fuel Rate (l/h)\\ 
			
			& & Absolute Load (\%)  \\ \cline{2-3}
			
			& \multicolumn{2}{c}{Short Term Fuel Trim B1 (\%)} \\

			& \multicolumn{2}{c}{Short Term Fuel Trim B2 (\%)}  \\

			& \multicolumn{2}{c}{Long Term Fuel Trim B1 (\%)}  \\

			& \multicolumn{2}{c}{Long Term Fuel Trim B2 (\%)}  \\
			\hline
			\textbf{Weather}& \multicolumn{2}{c}{Outside Air Temperature (\degree C)} \\
			\Xhline{2\arrayrulewidth}    
		\end{tabular}
	\end{table}

	\begin{table}[h!]
		\centering
		\caption{Static Data Attributes for VED}
		\label{Table: Static}
		\begin{tabular}{l l}
			\Xhline{2\arrayrulewidth}
			\textbf{Attribute} & \textbf{Example Values}\\      
			\Xhline{2\arrayrulewidth}
			Vehicle Type & ICE Vehicle, HEV, PHEV, EV \\
			\hline
			Vehicle Class & Passenger Car, SUV, Light Truck \\
			\hline
			Engine Configuration & I4, V4, V4 Flex, V6 PZEV \\
			\hline
			Engine Displacement & 1.0L, 2.0L, 3.6L \\
			\hline
			Transmission & 5-SP Automatic, 4-SP Manual, CVT \\
			\hline
			Drive Wheels & FWD, AWD \\
			\hline
			Vehicle Weights & 3,000lb, 5,000lb \\
			\Xhline{2\arrayrulewidth}

		\end{tabular}
	\end{table}

	\subsection{Data Preprocessing}
	
	\subsubsection{Energy efficiency estimation} 
	The four different vehicle types: ICE, HEV, PHEV, and EV utilize different types of energy: fuel-based energy, electric energy, or a combination of both. 
	A common measure of energy efficiency for ICE vehicles is mileage per gallon (MPG). The higher the mileage per gallon, the more efficient a vehicle is under given circumstances. For vehicles that utilize electric energy, an equivalent MPGe measure can be computed. Similarly, other measures such as L/100km can be utilized to estimate fuel energy efficiency. 
	
	For all trips, we first estimated the energy consumption. To compute the fuel consumption, we implemented the  algorithm described in~\ref{alg:fcr}, which is based on attributes available in the dataset. As for the battery energy, we computed the power in Watt based on the instantaneous values of current and voltage. Then, we integrated it over time to obtain the electric energy measure in kilowatt hour (kWh). We then calculated the average distance traveled per unit of energy consumed as a measure of energy efficiency to be able to compare the performance of different vehicles under different driving conditions. For the fuel energy, we considered as measure of energy efficiency the kilometer per liter (km/L) and for the electric battery energy, we considered the kilometer per kilowatt hour (km/kWh). 
	
	\begin{algorithm}[]
		\caption{Estimation of Fuel Consumption Rate (FCR)}
		\label{alg:fcr}
		\begin{algorithmic}
			\Require $FuelRate, MAF,AbsLoad, Displacement_{eng}$ \\
			\quad \quad $RPM_{eng}, STFT, LTFT, AFR, \rho_{air}$
			\Ensure $FCR$
			\State $ correction = (1 + STFT/100 + LTFT/100)/AFR $
			\If{$FuelRate$ is available}
			\State \Return $FuelRate$
			\ElsIf{$MAF$ is available}
			\State \Return $MAF \times correction$
			\ElsIf{$AbsLoad$ and $RPM_{eng}$ are available}
			\State $MAF = AbsLoad/100 \times \rho_{air} \times Displacement_{eng} \times RPM_{eng} / 120$
			\State \Return $MAF \times correction$
			\Else 
			\State  \Return NaN
			\EndIf
		\end{algorithmic}
	\end{algorithm}

	\subsubsection{Feature Engineering}
\begin{figure*}
		\centering
		\includegraphics[width=\textwidth]{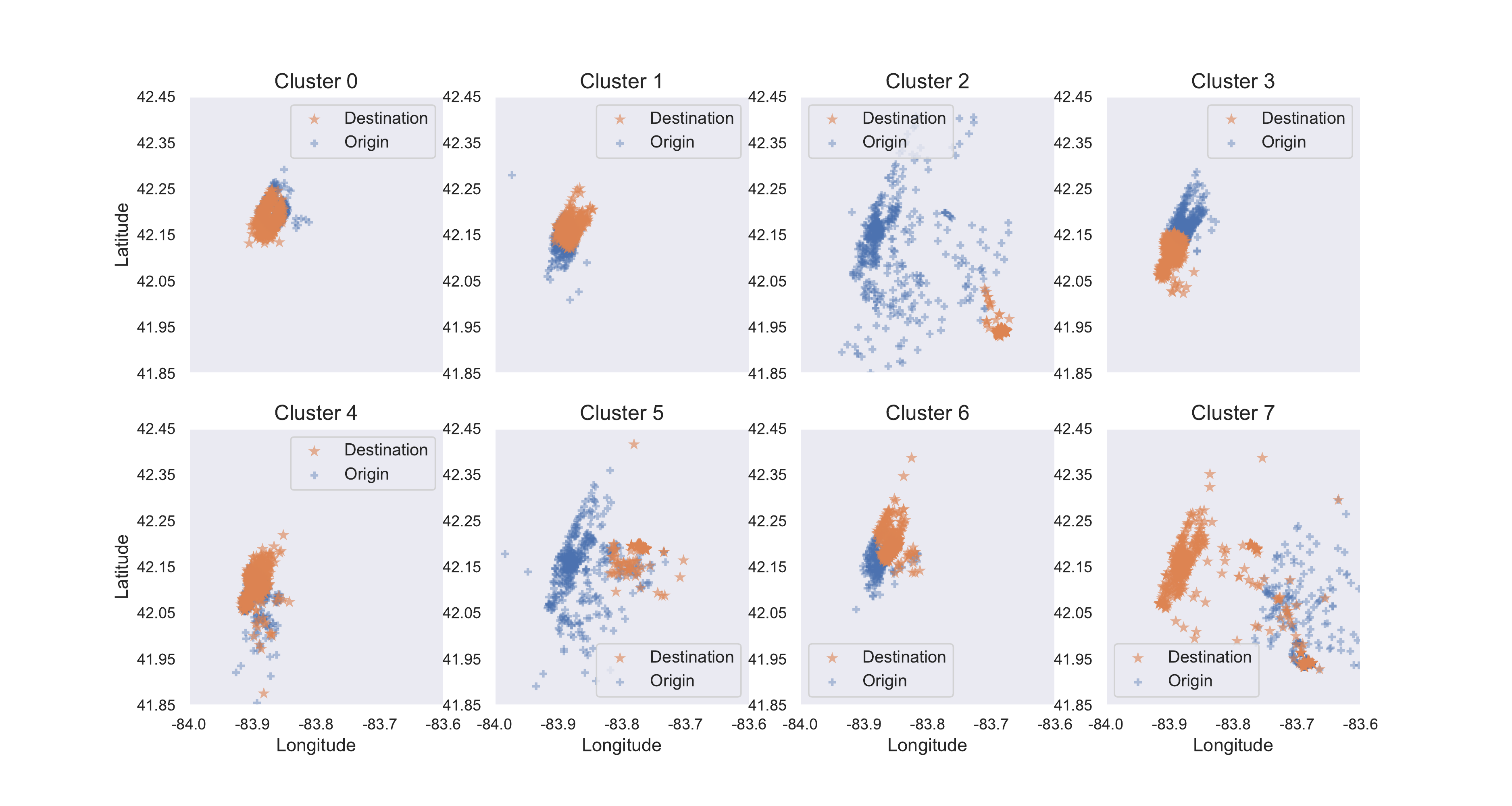}
		\caption{Origin-Destination Clustering for Trips}
		\label{fig:clusters}
	\end{figure*}

	In order to augment the dataset, we extracted relevant features from the existing ones. For instance, we generated time-based features from the existing timestamp: hour, minute, month, day of the week.
	Similarly, we added descriptive statistics about speed for each trip.
	Furthermore, we clustered the trips based on origins and destinations to group trips that took place in a similar geographical area or share a similar pattern. The cluster number was added as a feature of our dataset. Figure~\ref{fig:clusters} depicts the results of the trip clustering. For instance, we note that Cluster 2 illustrates trips that started and ended in a similar concentrated area, while Clusters 6 and 7 illustrate trips that span over a bigger geographical area. 
	
	\subsection{Ensemple Approach}
	Our ensemble approach is based on deep neural networks that are trained on stratified batches of the data, thus considering all months of the year.
	
	As shown in~\cite{lakshminarayanan2017simple}~\cite{shafahi2019adversarial}, we considered a dataset $D$, such as $D = \{x_n,y_n\}_{n=1}^N$, where $x \in \mathbb{R}^d$ stands for the d-dimensional features, and  $y \in \mathbb{R}$ is the prediction target. The base model, i.e. the neural network models the probabilistic predictive distribution $p_\theta(y|x)$, where $\theta$ are the parameters of the neural network. The method is based on three key aspects:
	
	\begin{enumerate}
		\item Using a scoring rule that reduces predictive uncertainty
		\item Using adversarial training to smooth the predictive distributions
		\item Training an ensemble of $M$ neural networks 
	\end{enumerate}
	
	As in~\cite{lakshminarayanan2017simple}, we used a network that delivered two values in the final layer: the predicted mean and variance, by treating the observed value as a sample from a Gaussian distribution. With the predicted mean and variance, we minimized the negative log-likelihood criterion expressed in Equation~\ref{eq:nll}.\\
	
	\begin{equation}
		\label{eq:nll}
		- \log p_{\theta}(y_n|x_n) = \frac{\log \sigma_\theta^2}{2}+ 
		\frac{(y-\mu_\theta(x))^2}{2\sigma_\theta^2(x)} + constant
	\end{equation}

	We trained an ensemble of neural networks independently in parallel. When training each network, we used random batches (subsamples) of the data at each iteration, as well as a random initialization of the parameters. 
	
	While the batches were drawn at random from the whole training set, the chronological order was later restored to take into account the original succession of the values. The resulting ensemble was a uniformly-weighted mixture model and combined the predictions as expressed in Equation~\ref{eq:ensemble}: 
	%$p(y|x)=M^{-1} \sum_{m=1}^{M} p_{\theta_m}(y|x,\theta_m)$. \\
	
	\begin{equation}
		\label{eq:ensemble}
		p(y|x)=M^{-1} \sum_{m=1}^{M} p_{\theta_m}(y|x,\theta_m)
	\end{equation}

	\section{Experiments and Results}
	\label{sec:results}

	\subsection{Experimental Setup}
	
	As a framework, we used Tensorflow~\cite{dillon2017tensorflow}. To build the ensemble, we used $M=10$ deep neural networks, a batch size of 500 and the Adam optimizer~\cite{zhang2018improved} with a fixed learning rate of 0.1 in our experiments. The individual neural networks are fully connected and 5 layers deep. We trained each neural network for 10 epochs. The results of the constituent networks are averaged in order to generate the ensemble's final prediction. A single neural network is represented in Figure~\ref{fig:nn_arch}, while Figure~\ref{fig:framework} depicts the ensemble structure.
	To tune the learning rate, we used a grid search on a log scale from 0.1 to $10^{-5}$. We performed a 70\%-30\% split between training and testing datasets in a stratified manner with respect to the month of the year, thus resulting in balanced training and test sets with respect to the chronological aspect of the data. Such a split allows to have a training and testing set that both include samples from all months of the year in a proportionate way, as opposed to a sequential split that would result in the testing set with samples from months that were not represented in the training set and vice versa.

	All experiments were performed on a computer with the following specifications:
	\begin{enumerate}
		\item \textbf{Processor}: Intel(R) Core(TM) i7 - 10510U CPU@1.8GHz 2.30 GHz
		\item \textbf{RAM}: 16 GB
		\item \textbf{GPU}: GeForce MX250 
	\end{enumerate}
	
	We opted to compare our findings to these baselines:
	\begin{enumerate}
		\item \textbf{Linear regression (LR)}~\cite{montgomery2021introduction}
		\item \textbf{Random Forest (RF)}~\cite{breiman2001random}
		\item \textbf{xgboost (XGB)}~\cite{chen2015xgboost}
		\item\textbf{A single neural network (NN)}~\cite{picton1994neural}
	\end{enumerate}
	
	The choice of these baselines is mainly motivated by their common use in regression tasks. In particular, linear regression is a widely used simple, fast, and easy to interpret method for regression tasks. As we are proposing an ensemble of deep neural networks, we deemed it relevant to compare our findings to commonly used ensembles such as random forest and xgboost. Both are based on decision trees and they leverage the diversity of their base models. To underline the added value of an ensemble, we also used the single neural network as a baseline, whereby it's trained following the same settings and parameters as the individual models of our ensemble.
	
	As an evaluation metric, we used the root mean square error (RMSE) which is computed over the same test set for our model and the baselines. This choice is justified by the prevalence of RMSE for evaluating regression tasks. For all models, we also considered the R2 score; which is a measure of goodness of fit of a given model to a regression task. It is obtained by computing the proportion of the variance in the dependent variable that is predictable from the independent variables.

	\subsection{Statistical Analysis}
	To determine if a statistically significant difference exists between the ensemble and the four baseline models, the Wilcoxon signed-rank test \cite{woolson2007wilcoxon} was conducted. It does not assume a distribution normality. Therefore, it is the non-parametric version of the paired T-test~\cite{hsu2007paired}. It allows to compare the errors generated by different predictive models and works by comparing the difference between two paired samples and ranking the absolute differences in magnitude. The null hypothesis for the Wilcoxon signed-rank test is that the median difference between the two paired samples is zero. The test statistic is calculated as the sum of the ranks of the positive differences or the sum of the ranks of the negative differences, whichever is smaller. 
	
	More specifically, to determine whether one model performs significantly better than the others, we performed the one-tailed hypothesis test. The null hypothesis ($H_0$) is that “Two models have the same predictive error”. The alternative hypothesis ($H_a$) is that “The first model has a smaller error than the second model”. In the Wilcoxon signed-rank test, we chose a significance level of $0.05$. In this case, the Wilcoxon signed-rank test is run using a dedicated function of the \texttt{Python} package \texttt{scipy}. The test was conducted to compare the ensemble to each baseline.
	
	\subsection{Results}
\begin{figure*}
		\centering
		\begin{subfigure}{0.28\textwidth}
			\includegraphics[width=\textwidth]{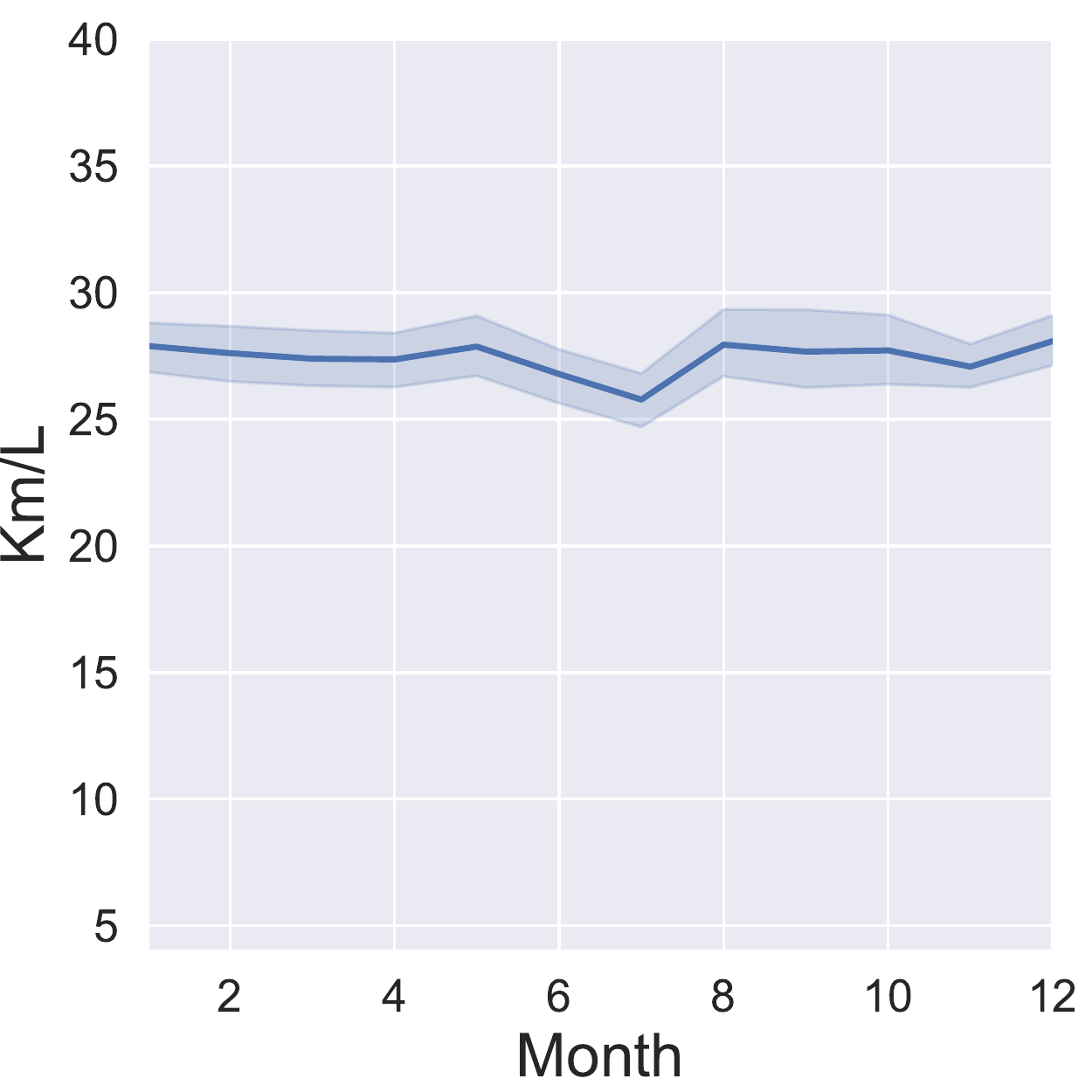}
			\caption{ICE}
			\label{fig:ice_fuel}
		\end{subfigure}
		\hfill
		\begin{subfigure}{0.28\textwidth}
			\includegraphics[width=\textwidth]{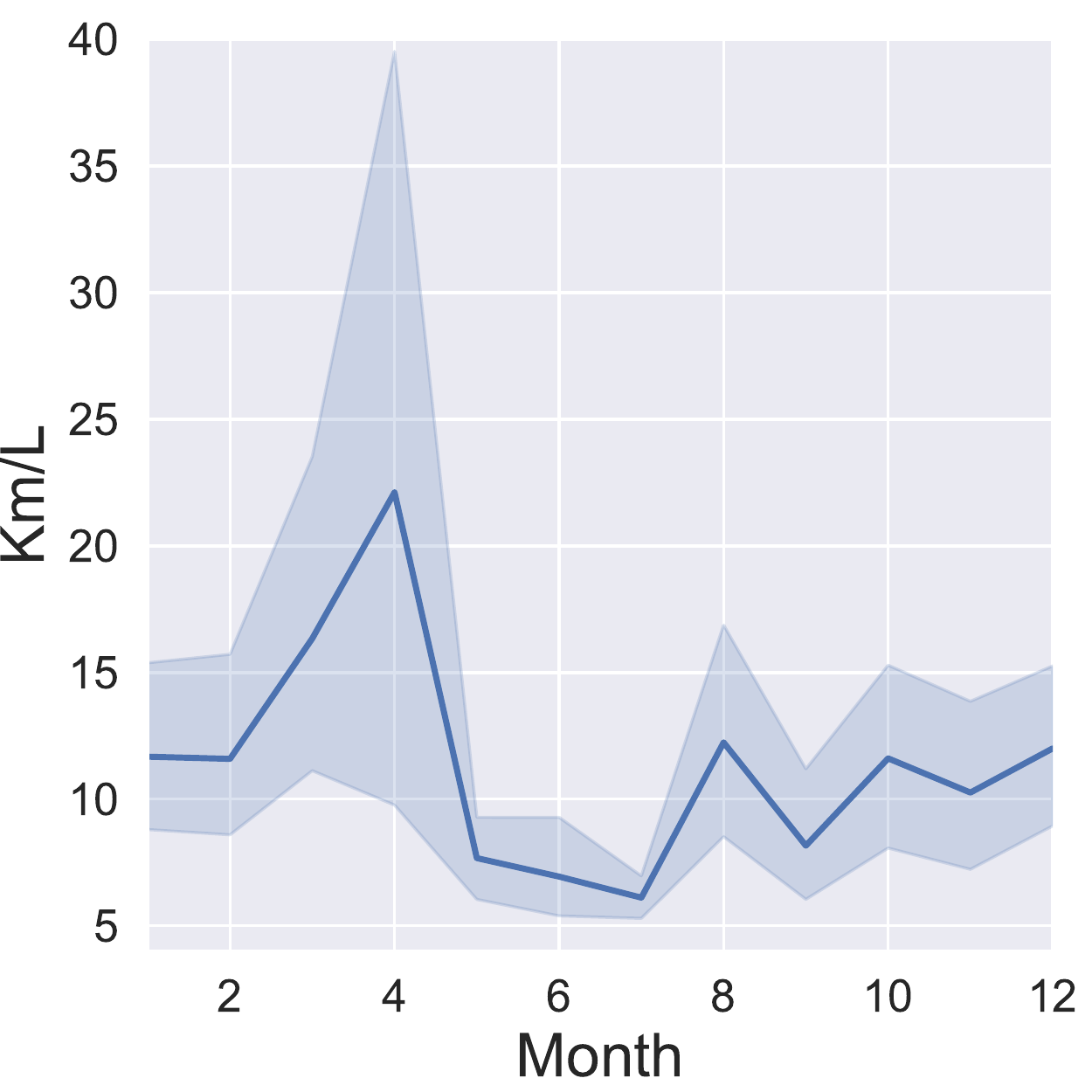}
			\caption{PHEV}
			\label{fig:phev_fuel}
		\end{subfigure}
		\hfill
		\begin{subfigure}{0.28\textwidth}
			\includegraphics[width=\textwidth]{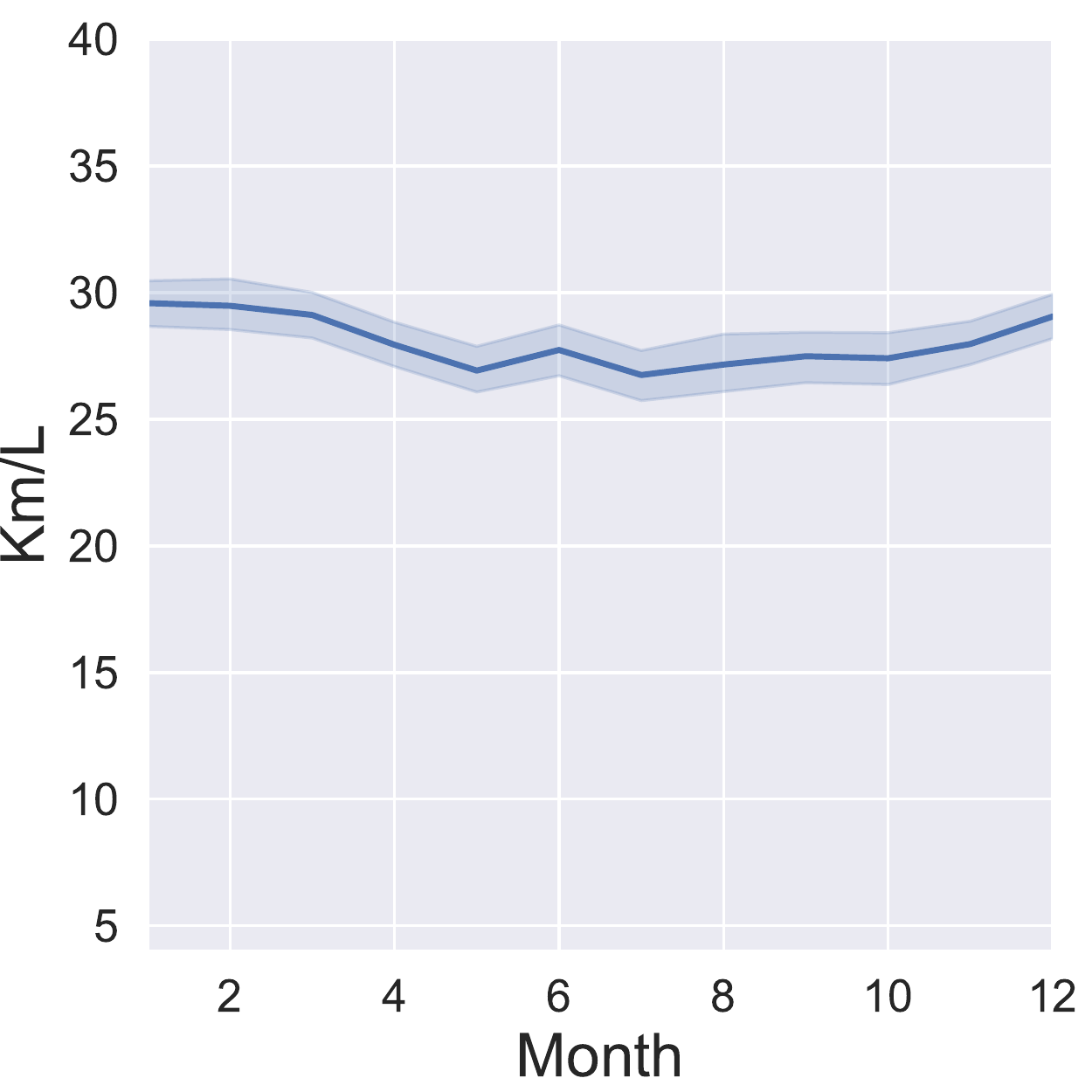}
			\caption{HEV}
			\label{fig:hev_fuel}
		\end{subfigure}
		
		\begin{subfigure}{0.28\textwidth}
			\includegraphics[width=\textwidth]{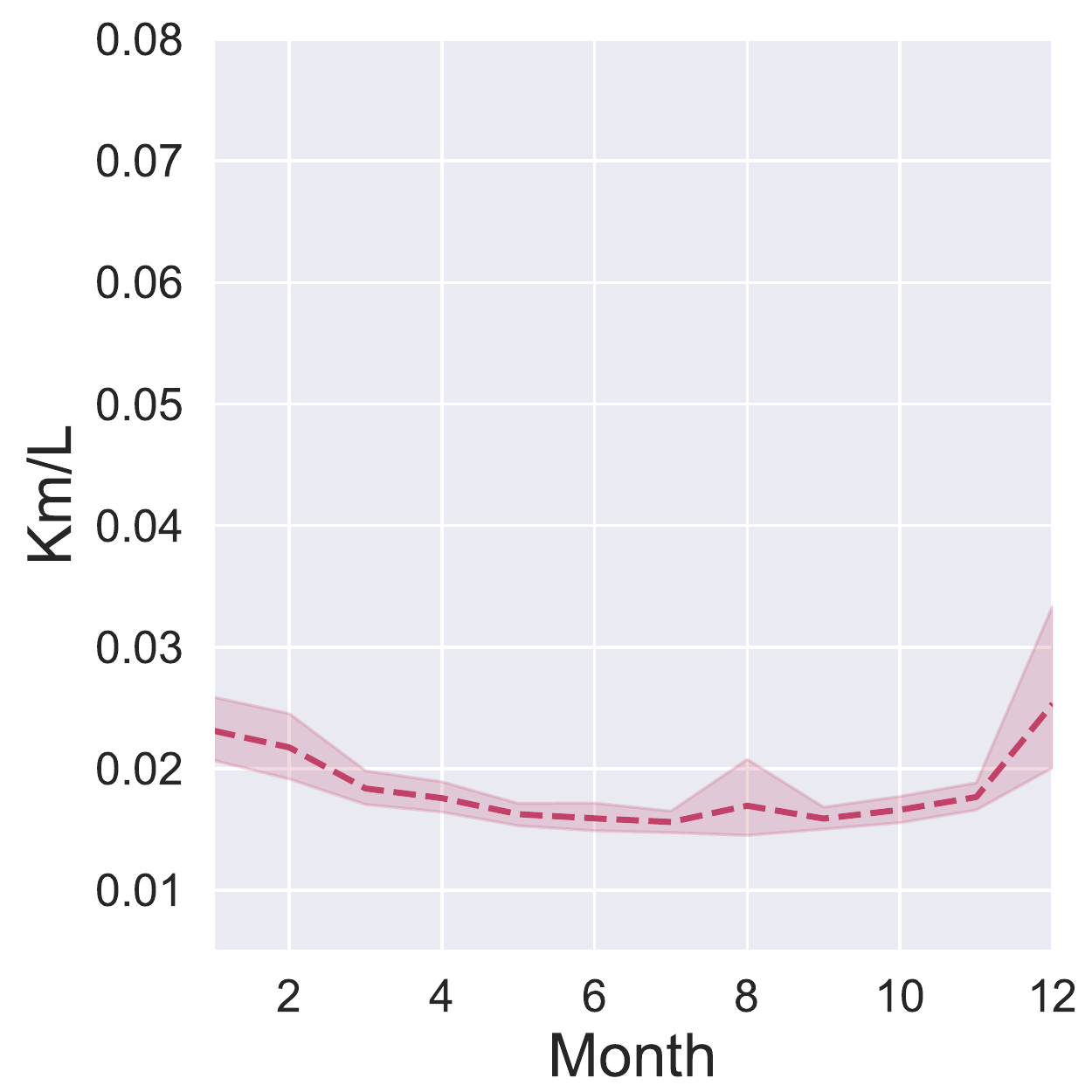}
			\caption{PHEV}
			\label{fig:phev_bat}
		\end{subfigure}
		\hfill
		\begin{subfigure}{0.28\textwidth}
			\includegraphics[width=\textwidth]{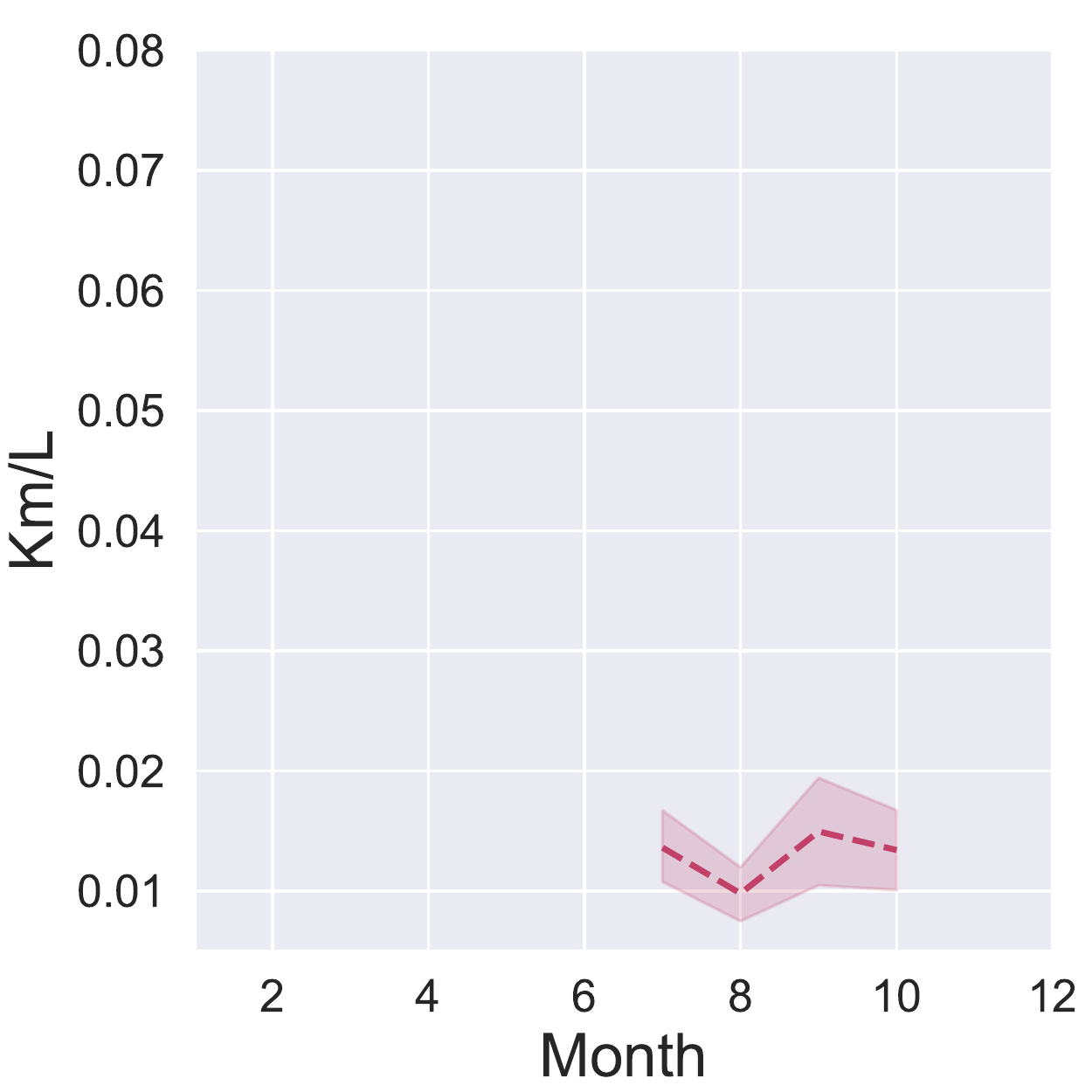}
			\caption{HEV}
			\label{fig:hev_bat}
		\end{subfigure}
		\hfill
		\begin{subfigure}{0.28\textwidth}
			\includegraphics[width=\textwidth]{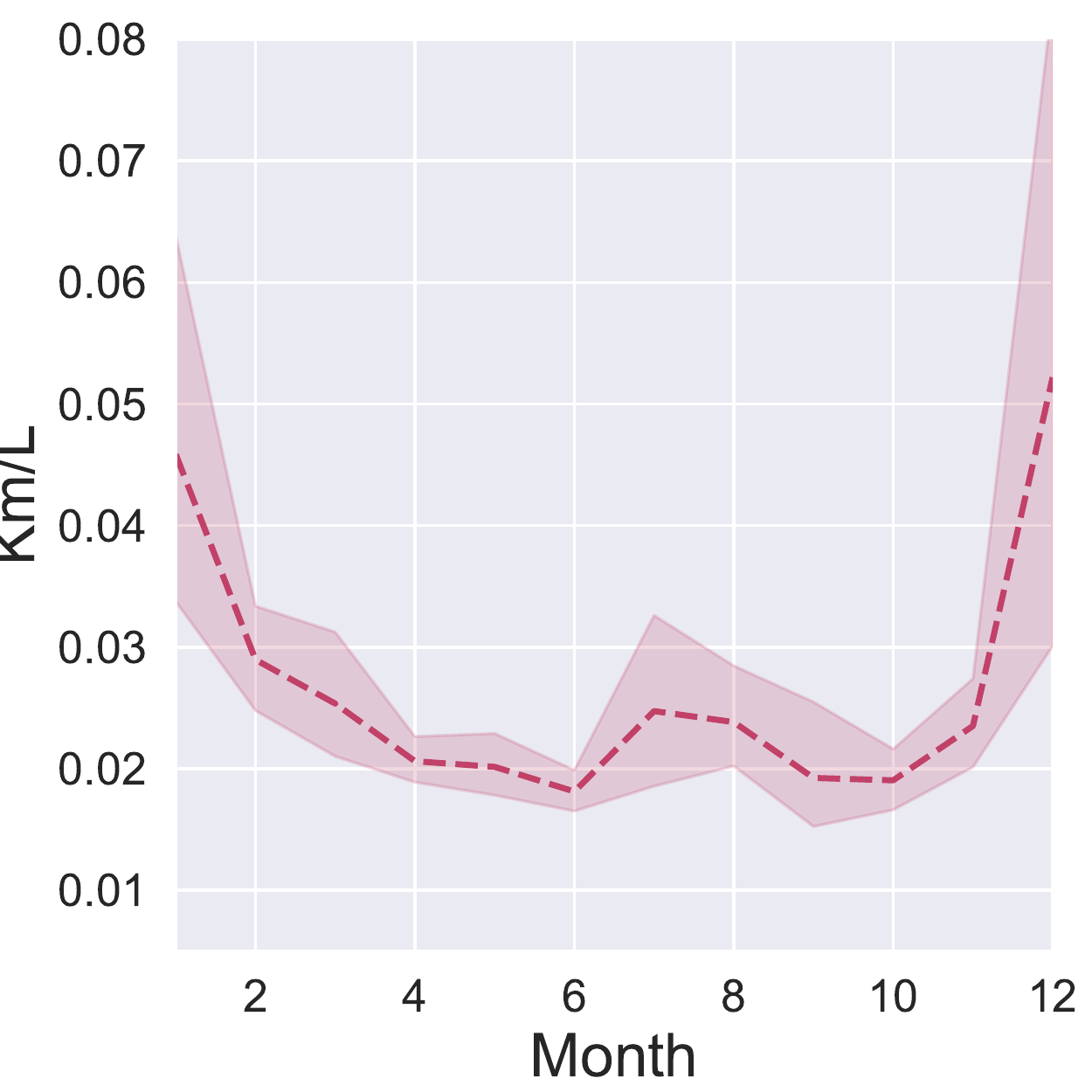}
			\caption{EV}
			\label{fig:ev_bat}
		\end{subfigure}

		\caption{Ensemble predictions of energy efficiency per vehicle, per energy type, and per month. Expressed in blue and solid lines are the fuel-related energy efficiency values, in red and dashed lines the values from the electric-based energy.} 
		\label{fig:results}
	\end{figure*}

	Figure~\ref{fig:results} illustrates the prediction results of our ensemble as well as the measure of the predictive uncertainty. For each plot, the line represents the mean values of predictions, while the shadow in a lighter color represents the uncertainty measure. The higher the span of the variance, i.e. the shadow in the figure, the higher the uncertainty. On the contrary, when the variance span is small, it indicates a lower uncertainty and hence a higher confidence in the prediction. 
	
	As the available data span over a period of a year, we report the results over the testing set per month, to highlight the temporal distribution of the results. This is enabled by our stratified approach in the training and testing split, thus allowing us to show test results for all months of the year.
	
	In blue, Figures~\ref{fig:ice_fuel}, \ref{fig:phev_fuel}, and ~\ref{fig:hev_fuel} show the fuel efficiency prediction results for ICE, PHEV, and HEV vehicles, respectively. The fuel efficiency predictions are reported in terms of Km/L i.e. traveled distance per unit of fuel consumption. 
	As for the figures in red; \ref{fig:phev_bat}, \ref{fig:hev_bat}, and ~\ref{fig:ev_bat} show the prediction results for the electric battery efficiency for PHEV, HEV, and EV vehicles, respectively. The battery efficiency predictions are reported in terms of Km/kWh i.e. traveled distance per unit of electric energy consumption. Therefore, for both types of energy, the higher the predicted values, the higher the efficiency. Due to lack of data for HEV vehicles, Figure~\ref{fig:hev_bat} shows prediction results only for four months: July to October. As for Figures~\ref{fig:phev_bat} and~\ref{fig:ev_bat}, the available data covers the whole year.

	As detailed in Section~\ref{sec:method}, we compared the performance of our ensemble ENN to that of several baselines: LR, RF, XGB, and NN. The computed results are reported per vehicle and energy type in Table ~\ref{table:results}. The values in the table are in terms of root mean square error (RMSE) calculated over the same test set.

	Additionally, we report in Table~\ref{table:r2_results} the R2 values for our ensemble as well as the baselines. We note that the R2 scores are consistent with the RMSE values in~\ref{table:results}. Therefore, the models with the highest R2 scores are also the models that perform better in the regression task. This is expected as the R2 score reflects the goodness of fit of a given model to a regression task. The closer the R2 score is to 1, the better the goodness of fit. Our ensemble has R2 scores between 0.74 and 0.92 for the six different prediction tasks per vehicle type and per energy type. Similarly to the RMSE values, we note a higher R2 score for xgboost for PHEV and EV battery energy efficiency.
	
	As for the statistical test, the results of the Wilcoxon signed-rank test are summarized in Table~\ref{table:p_results}. Each row details the results for the ensemble compared to one baseline. For the fuel efficiency prediction tasks, we note that the ensemble performs significantly better than other baselines for all vehicle types (ICE, PHEV, and HEV), since the p-value is lower than 0.05. As for the battery efficiency, the gain in performance is mostly not as significant. For instance, the ensemble does not perform significantly better than any of the baselines in the case of EV and HEV battery efficiency prediction. This effect can be attributed to the lack of data available for EV and HEV vehicles, resulting in a gain in performance that is not significant. In the case of battery efficiency prediction for PHEV, the ensemble performs significantly better compared to LR, RF, and XGB, but not NN. Overall, considering the distribution of trips per vehicle type, we note that the ensemble provides a considerable gain in performance in most cases.
	
	\begin{table*}[]
		\centering
		\caption{Prediction Results for Ensemble and Baselines. The values are expressed as mean $\pm$ standard  deviation, and reported in terms of root mean square error (RMSE). The lowest errors are in bold. }
		\label{table:results}
		\begin{adjustbox}{width=0.9\textwidth}
			\begin{tabular}{l?lll?lll}
				\cline{2-7}
				\Xhline{2\arrayrulewidth}
				& \multicolumn{3}{c?}{\textbf{Fuel Efficiency (RMSE)}}                                                                     & \multicolumn{3}{c}{\textbf{Battery Efficiency (RMSE)}}                                                                   \\ \hline
				
				& \multicolumn{1}{c|}{\textbf{ICE}}    & \multicolumn{1}{c|}{\textbf{PHEV}}   & \multicolumn{1}{c?}{\textbf{HEV}} & \multicolumn{1}{c|}{\textbf{HEV}}    & \multicolumn{1}{c|}{\textbf{PHEV}}    & \multicolumn{1}{c}{\textbf{EV}} \\ 
				\Xhline{2\arrayrulewidth}
				\textbf{LR} & \multicolumn{1}{l|}{$6.54 \pm 1.95$} & \multicolumn{1}{l|}{$17.39 \pm 4.14$} & $5.5 \pm 1.75$ & \multicolumn{1}{l|}{$0.025 \pm 0.01$}   & \multicolumn{1}{l|}{$0.031 \pm 0.015$}  & $0.039 \pm 0.011$  \\ \hline
				\textbf{RF}     & \multicolumn{1}{l|}{$5.22 \pm 0.87$}  & \multicolumn{1}{l|}{$15.72 \pm 3.28$}  & $4.01 \pm 2.1$  & \multicolumn{1}{l|}{$0.014 \pm 0.006$}   & \multicolumn{1}{l|}{$0.018 \pm 0.003$}  & $0.019 \pm 0.002$  \\ \hline
				\textbf{XGB}           & \multicolumn{1}{l|}{$2.92 \pm 0.93$}  & \multicolumn{1}{l|}{$12.28 \pm 3.36$}  & $4.9 \pm 2.35$  & \multicolumn{1}{l|}{$0.015 \pm 0.005$}   & \multicolumn{1}{l|}{\textbf{0.008 $\pm$ 0.004}}  & \textbf{0.015 $\pm$ 0.009}  \\ \hline
				\textbf{NN}        & \multicolumn{1}{l|}{$3.01 \pm 1.01$} & \multicolumn{1}{l|}{$9.45 \pm 4.12$}  & $3.45 \pm 1.02$  & \multicolumn{1}{l|}{$0.007 \pm 0.002$}   & \multicolumn{1}{l|}{$0.013 \pm 0.003$}  & $0.024 \pm 0.007$  \\ \hline
				\textbf{ENN}          & \multicolumn{1}{l|}{\textbf{2.45 $\pm$ 0.85}} & \multicolumn{1}{l|}{\textbf{6.79 $\pm$ 2.14}} & \textbf{2.87 $\pm$ 0.52} & \multicolumn{1}{l|}{\textbf{0.005 $\pm$ 0.002}}   & \multicolumn{1}{l|}{$0.01 \pm 0.002$}  & $0.021 \pm 0.003$ \\ \Xhline{2\arrayrulewidth} 
			\end{tabular}
		\end{adjustbox}
	\end{table*}

	\begin{table*}[!htb]
		%\caption{Global caption}
		\begin{minipage}{.5\linewidth}
			\caption{R2 Scores; the highest values are in bold.}
			\label{table:r2_results}
			\centering
			\begin{tabular}{l?lll?lll}
				\cline{2-7}
				\Xhline{2\arrayrulewidth}
				& \multicolumn{3}{c?}{\textbf{Fuel Efficiency (R2)}}                                                                     & \multicolumn{3}{c}{\textbf{Battery Efficiency (R2)}}                                                                   \\ \hline
				
				& \multicolumn{1}{c|}{\textbf{ICE}}    & \multicolumn{1}{c|}{\textbf{PHEV}}   & \multicolumn{1}{c?}{\textbf{HEV}} & \multicolumn{1}{c|}{\textbf{HEV}}    & \multicolumn{1}{c|}{\textbf{PHEV}}    & \multicolumn{1}{c}{\textbf{EV}} \\ 
				\Xhline{2\arrayrulewidth}
				\textbf{LR}  & \multicolumn{1}{l|}{$0.68$}  & \multicolumn{1}{l|}{$0.72$}  & $0.73$  & \multicolumn{1}{l|}{$0.65$}   & \multicolumn{1}{l|}{$0.62$}  & $0.7$  \\ \hline
				\textbf{RF}     & \multicolumn{1}{l|}{$0.83$}  & \multicolumn{1}{l|}{$0.77$}  & $0.89$  & \multicolumn{1}{l|}{$0.76$}   & \multicolumn{1}{l|}{$0.77$}  & $0.65$  \\ \hline
				\textbf{XGB}           & \multicolumn{1}{l|}{$0.8$}  & \multicolumn{1}{l|}{$0.82$}  & $0.87$  & \multicolumn{1}{l|}{$0.73$}   & \multicolumn{1}{l|}{\textbf{0.82}}  & \textbf{0.78}  \\ \hline
				\textbf{NN}        & \multicolumn{1}{l|}{$0.85$} & \multicolumn{1}{l|}{$0.79$}  & $0.9$  & \multicolumn{1}{l|}{$0.83$}   & \multicolumn{1}{l|}{$0.74$}  & $0.72$  \\ \hline
				\textbf{ENN}          & \multicolumn{1}{l|}{\textbf{0.92}} & \multicolumn{1}{l|}{\textbf{0.86}} & \textbf{0.94} & \multicolumn{1}{l|}{\textbf{0.87}}   & \multicolumn{1}{l|}{$0.78$}  & $0.74$ \\ \Xhline{2\arrayrulewidth} 
			\end{tabular}
		\end{minipage}%
		\begin{minipage}{.5\linewidth}
			\centering
			\caption{The results of the Wilcoxon signed-rank test.}
			\label{table:p_results}
			\begin{tabular}{l?lll?lll}
				\cline{2-7}
				\Xhline{2\arrayrulewidth}
				& \multicolumn{3}{c?}{\textbf{Fuel Efficiency (\textit{P}-value)}}                                                                     & \multicolumn{3}{c}{\textbf{Battery Efficiency (\textit{P}-value)}}                                                                   \\ \hline
				
				& \multicolumn{1}{c|}{\textbf{ICE}}    & \multicolumn{1}{c|}{\textbf{PHEV}}   & \multicolumn{1}{c?}{\textbf{HEV}} & \multicolumn{1}{c|}{\textbf{HEV}}    & \multicolumn{1}{c|}{\textbf{PHEV}}    & \multicolumn{1}{c}{\textbf{EV}} \\ 
				\Xhline{2\arrayrulewidth}
				\textbf{ENN-LR}  & \multicolumn{1}{l|}{$0.022$}  & \multicolumn{1}{l|}{$0.042$}  & $0.015$  & \multicolumn{1}{l|}{$0.035$}   & \multicolumn{1}{l|}{$0.017$}  & $0.053$  \\ \hline
				\textbf{ENN-RF}     & \multicolumn{1}{l|}{$0.031$}  & \multicolumn{1}{l|}{$0.033$}  & $0.035$  & \multicolumn{1}{l|}{$0.053$}   & \multicolumn{1}{l|}{$0.036$}  & $0.079$  \\ \hline
				\textbf{ENN-XGB}           & \multicolumn{1}{l|}{$0.037$}  & \multicolumn{1}{l|}{$0.041$}  & $0.033$  & \multicolumn{1}{l|}{$0.076$}   & \multicolumn{1}{l|}{0.039}  & $0.064$  \\ \hline
				\textbf{ENN-NN}        & \multicolumn{1}{l|}{$0.028$} & \multicolumn{1}{l|}{$0.039$}  & $0.035$  & \multicolumn{1}{l|}{$0.062$}   & \multicolumn{1}{l|}{$0.057$}  & $0.084$ \\ \Xhline{2\arrayrulewidth} 
			\end{tabular}
		\end{minipage} 
	\end{table*}
	
	%\begin{figure*}
	%   \centering
	%   \begin{subfigure}{0.32\textwidth}
		%       \includegraphics[width=\textwidth]{bar_ice_fuel}
		%       \caption{ICE}
		%       \label{fig:bar_ice_fuel}
		%   \end{subfigure}
	%   \hfill
	%   \begin{subfigure}{0.32\textwidth}
		%       \includegraphics[width=\textwidth]{bar_phev_fuel}
		%       \caption{PHEV}
		%       \label{fig:bar_phev_fuel}
		%   \end{subfigure}
	%   \hfill
	%   \begin{subfigure}{0.32\textwidth}
		%       \includegraphics[width=\textwidth]{bar_hev_fuel}
		%       \caption{HEV}
		%       \label{fig:bar_hev_fuel}
		%   \end{subfigure}
	%   
	%   \begin{subfigure}{0.32\textwidth}
		%       \includegraphics[width=\textwidth]{bar_phev_bat}
		%       \caption{PHEV}
		%       \label{fig:bar_phev_bat}
		%   \end{subfigure}
	%   \hfill
	%   \begin{subfigure}{0.32\textwidth}
		%       \includegraphics[width=\textwidth]{bar_hev_bat}
		%       \caption{HEV}
		%       \label{fig:bar_hev_bat}
		%   \end{subfigure}
	%   \hfill
	%   \begin{subfigure}{0.32\textwidth}
		%       \includegraphics[width=\textwidth]{bar_ev_bat}
		%       \caption{EV}
		%       \label{fig:bar_ev_bat}
		%   \end{subfigure}
	%   
	%   
	%   \caption{RMSE error barplots comparing models per vehicle type} 
	%   \label{fig:results_bar}
	%\end{figure*}

	\section{Discussion}
	\label{sec:discussion}
	
	We note that for figures ~\ref{fig:ice_fuel} and~\ref{fig:hev_fuel} relative to ICE and HEV vehicles show higher values of mean fuel efficiency predictions as well as smaller uncertainty ranges. On the other hand, the results for PHEV vehicles shown in Figure ~\ref{fig:phev_fuel} demonstrate more fluctuations of the fuel efficiency, as well as a higher uncertainty ranges, which indicate that the underlying data have more variance and randomness. Therefore, the model cannot produce predictions with high confidence.

	We note that in the colder months (January, November, December) the uncertainty range as well as the mean predicted efficiency shows a distinct increase compared to the rest of the year. This may interpreted as a result of varying weather conditions such as rainy, snowy, and dry weather, which in return affect the road conditions and driving styles. Although the data that we used does not identify the drivers, we note that the data spans over a year and stems from 383 vehicles. Therefore, it includes data from different drivers. Consequently, the data we used implicitly includes variance in driver behavior and driving conditions. Furthermore, we note that for all three vehicle types that are equipped with batteries, the uncertainty range increases between July and August. This effect possibly reflects the use of air conditioning which increases the use of energy and introduces more variance in the data. 
	
	Table~\ref{table:results} shows the prediction error results in terms of RMSE for the various considered baselines. The linear regression consistently has the highest error values, likely due to the low complexity of the model compared to the complexity of the data. The ensemble results for the different use cases outperform the single neural net's. Additionally, the standard deviations of the RMSE errors of our neural networks ensemble are consistently smaller than those of the baselines, thus indicating a more consistent performance across the test samples. As for XGB, we note that it only outperforms the ensemble in the case of HEV and EV battery efficiency prediction. This is likely due to the reduced amount of data available for training in these particular cases compared to the remaining ones. As noted in the data description, there are only 27 PHEV and EV vehicles compared to 264 ICE and 92 HEV vehicles that are represented in the data. In all other cases, ENN has the lowest error values, thus demonstrating a high predictive accuracy compared to baselines. Consequently, we note that DNNs in general, and the DNN ensemble in particular require a high amount of data for sufficient training. On the other hand, tree-based ensembles such as XGB can be less sensitive to a lack of data and this gives them an advantage compared to more complex models.

	Our proposed approach is capable of producing accurate predictions of the energy efficiency as shown in terms of RMSE and R2 score. Furthermore, the statistical test showed a significantly higher performance than baselines in most cases. The consistent performance across different types of vehicles and types of used energy indicates the versatility of the ensemble approach.  In addition, the output of a measure of uncertainty as shown in Figure~\ref{fig:results} provides further information about the model's confidence in the predictions, which is beneficial due to the variance present in the data and the multiple sources of uncertainty: weather conditions, driving behavior, type and state of vehicle, etc.

	A possible limitation of using neural network ensemble is the increased initial training time compared to other ensembles such as decision-tree based ensembles. However, fast predictions can be provided in test time, thus facilitating deployment. On the other hand, the consistent performance across the training and testing sets indicate that the ensemble has not overfit to the training data. The considered data which span over a year is therefore deemed sufficient.
	
	All in all, this study demonstrates an approach for predicting energy efficiency by relying only on floating car and sensor data. Our findings are in agreement with works in the literature that highlight how energy efficiency varies according to multiple factors such as the type of vehicle, driving conditions, weather, etc. On the other hand, by relying only on floating car and sensor data, we are able to propose an approach that can be deployed in different applications such as fleet management and trip planning, where providing the uncertainty of the estimations can be essential for decision-making, user trust, and energy saving policies.

	\section{Conclusion}
	\label{sec:conclusion}
	The high use of energy in the field of transportation can be considerably reduced by increasing the energy efficiency. In fact, predicting energy efficiency is relevant because it provides a comprehensive assessment, which is useful for evaluating the effectiveness of energy-saving technologies and policies, as well as for making informed decisions about vehicle purchases and usage. To do so, we proposed an approach based on an ensemble of neural networks (ENN). While most available works in the literature focus on one type of vehicle,  we relied on publicly available energy data (VED) which include different vehicle types over a period of a year. Our approach ensured a high predictive accuracy, but also provided information about the model's confidence in its output, thus facilitating decision making and increasing trust in the model's output. 
	Future work will focus on investigating various measures of uncertainty and uncoupling aleatoric and epistemic uncertainty in this particular context.

	\section*{Acknowledgment}
	This work was supported by the Austrian Ministry for Climate Action, Environment, Energy, Mobility, Innovation and Technology (BMK) Endowed Professorship for Sustainable Transport Logistics 4.0., IAV France S.A.S.U., IAV GmbH, Austrian Post AG and the UAS Technikum Wien. 
	
	\bibliographystyle{IEEEtran}
	\bibliography{jk_jabref}
	
	\begin{IEEEbiography}[{\includegraphics[width=1in,height=1.25in,clip,keepaspectratio]{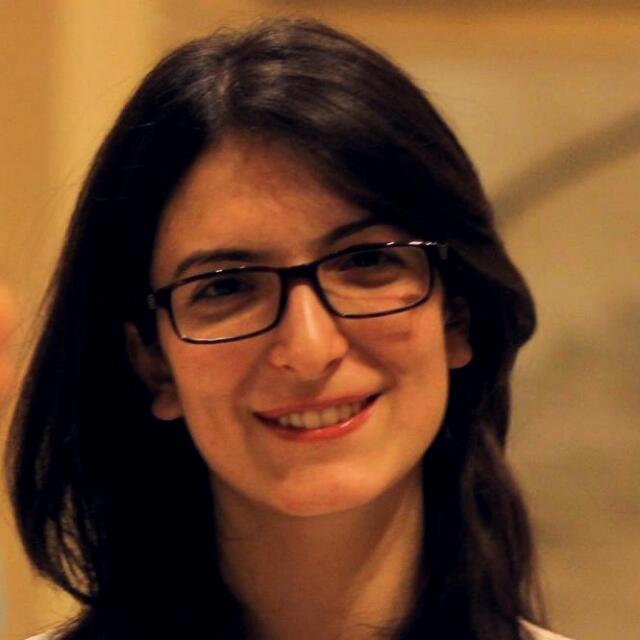}}]{Jihed Khiari}  graduated with a Master's degree in Computer Science with a specialization in Machine Learning from the National School of Computer Science in Tunisia. She is now a PhD student and a research assistant at the Johannes Kepler University (JKU) Linz in Austria. She works under the supervision of Prof. Dr. Olaverri-Monreal at the ITS-Chair for Sustainable Transport Logistics 4.0. Her research interests include applying machine learning techniques to optimize transportation systems and road traffic. Before joining JKU, she worked as a research scientist at a research and development company, where she contributed to several EU and internal projects related to intelligent transportation systems.
	\end{IEEEbiography}

	\begin{IEEEbiography}[{\includegraphics[width=1in,height=1.25in,clip,keepaspectratio]{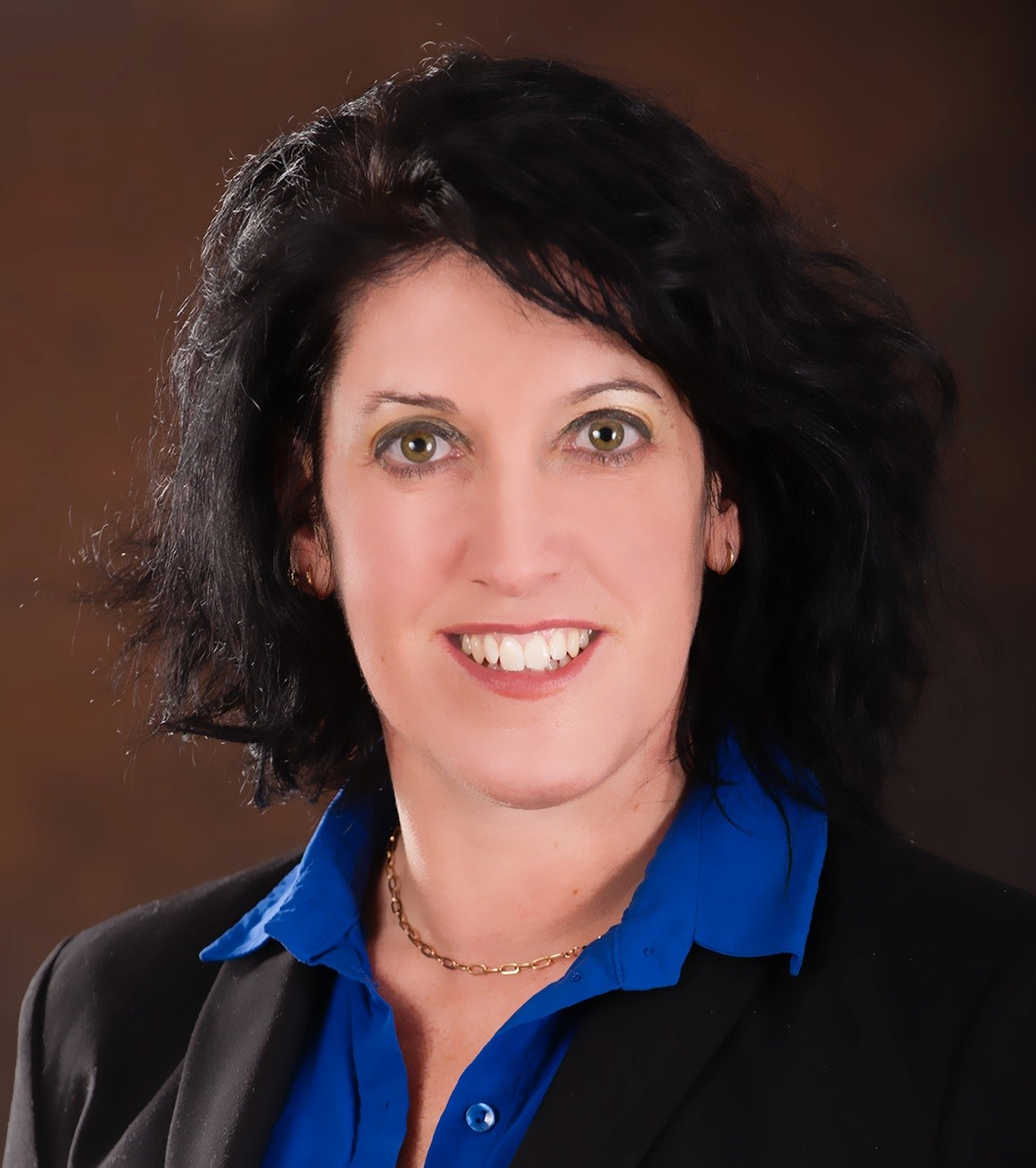}}]{Univ. Prof. Dr. Cristina Olaverri-Monreal} is professor and holder of the BMK endowed chair Sustainable Transport Logistics 4.0 at the Johannes Kepler University Linz, in Austria. Prior to this position, she led diverse teams in the industry and in the academia in the US and in distinct countries in Europe.
		She is the president of the IEEE Intelligent Transportation Systems Society (IEEE ITSS), founder and chair of the Austrian IEEE ITSS chapter, and chair of the Technical Activities Committee (TAC) on Human Factors in ITS.        
		She received her PhD from the Ludwig-Maximilians University (LMU) in Munich in cooperation with BMW. Her research aims at studying solutions for an efficient and effective transportation focusing on minimizing the barrier between users and road systems. To this end, she relies on the automation, wireless communication and sensing technologies that pertain to the field of Intelligent Transportation Systems (ITS).
		Dr. Olaverri is a member of the EU-wide platform for coordinating open road tests (Cooperative, Connected and Automated Mobility (CCAM)) as well as a representative for the European technology platform "Alliance for Logistics Innovation through Collaboration in Europe" (ALICE) for the "Workgroup Road Safety" (WG4: EU-CCAM-WG-ROAD-SAFETY@ec.europa.eu). She is additionally a senior/associate editor and editorial board member of several journals in the field, including the IEEE ITS Transactions and IEEE ITS Magazine. Furthermore, she is an expert for the European Commission on "Automated Road Transport" and consultant and project evaluator in the field of ICT and "Connected, Cooperative Autonomous Mobility Systems" for various EU and national agencies as well as organizations in Germany, Sweden, France, Ireland, etc. In 2017, she was the general chair of the "IEEE International Conference on Vehicles Electronics and Safety" (ICVES'2017). She was awarded the "IEEE Educational Activities Board Meritorious Achievement Award in Continuing Education" for her dedicated contribution to continuing education in the field of ITS. 
	\end{IEEEbiography}

\end{document}